\documentclass[runningheads]{llncs}
\usepackage[T1]{fontenc}
\usepackage{graphicx}

\usepackage{newfloat}
\usepackage{listings}
\usepackage{threeparttable} 
\usepackage{booktabs,makecell,multirow} 
\usepackage{color, soul, colortbl}
\usepackage[dvipsnames]{xcolor}
\usepackage{verbatim}
\usepackage{amsfonts, amssymb, amsmath, bm}
\usepackage{subfigure}
\usepackage{tabularx}
\usepackage{float}
\usepackage{marvosym}

\begin{document}
%
\title{XCoOp: Explainable Prompt Learning for Computer-Aided Diagnosis via Concept-guided Context Optimization}
%
%
\author{Yequan Bie\inst{1} \and
Luyang Luo\inst{1}\and
Zhixuan Chen\inst{1}\and 
Hao Chen\inst{1,2}\textsuperscript{\Letter}}

%
\authorrunning{Y. Bie et al.}
%

\institute{
    \textsuperscript{\rm 1}Department of Computer Science and Engineering, Hong Kong University of Science and Technology, Hong Kong, China\\
    \textsuperscript{\rm 2}Department of Chemical and Biological Engineering, Hong Kong University of Science and Technology, Hong Kong, China\\
\email{\{ybie, zchenhi\}@connect.ust.hk, cseluyang@ust.hk, jhc@cse.ust.hk}
}
%
\maketitle              

\newcommand\blfootnote[1]{%
\begingroup
\renewcommand\thefootnote{}\footnote{#1}%
\addtocounter{footnote}{-1}%
\endgroup
}

\blfootnote{\Letter \, Corresponding author.}

\begin{abstract}
Utilizing potent representations of the large vision-language models (VLMs) to accomplish various downstream tasks has attracted increasing attention. Within this research field, soft prompt learning has become a representative approach for efficiently adapting VLMs such as CLIP, to tasks like image classification. However, most existing prompt learning methods learn text tokens that are unexplainable, which cannot satisfy the stringent interpretability requirements of Explainable Artificial Intelligence (XAI) in high-stakes scenarios like healthcare. To address this issue, we propose a novel explainable prompt learning framework that leverages medical knowledge by aligning the semantics of images, learnable prompts, and clinical concept-driven prompts at multiple granularities. Moreover, our framework addresses the lack of valuable concept annotations by eliciting knowledge from large language models and offers both visual and textual explanations for the prompts. Extensive experiments and explainability analyses conducted on various datasets, with and without concept labels, demonstrate that our method simultaneously achieves superior diagnostic performance, flexibility, and interpretability, shedding light on the effectiveness of foundation models in facilitating XAI. The code will be made publically available.

\end{abstract}
\section{Introduction}

In the era of foundation models (FMs), large-scale vision-language pre-trained models (VLMs) such as CLIP \cite{clip}, BLIP \cite{li2022blip}, Flamingo \cite{alayrac2022flamingo}, ALIGN \cite{jia2021align}, CoCa \cite{yu2205coca} have underscored their potential in representation learning, excelling at vision and language understanding. However, the massive sizes and expensive training costs have prompted studies to explore ways to efficiently adapt the knowledge of pre-trained VLMs to downstream tasks. Recently, prompt learning from the field of natural language processing has been introduced to the vision domain \cite{coop,cocoop}, achieving great success in adapting large-scale VLMs to downstream tasks like image classification and segmentation \cite{segment,seg2}. These methods fix the parameters of the models and train the learnable tokens that serve as the input of the text encoder (i.e., context optimization), significantly reducing the cost of utilizing foundation models. Nevertheless, 
existing prompt learning methods result in unexplainable learned tokens. 
This lack of interpretability prevents further application of prompt learning from 
being applied to high-stakes domains with rigorous demands of trustworthiness, such as healthcare \cite{lipton2017doctor,rudin2019stop}. Specifically, the models applied to the healthcare domain should not only perform well but also need to be understandable and trustworthy to practitioners, necessitating research into Explainable Artificial Intelligence (XAI). Several prior studies have introduced knowledge to prompt learning \cite{kgcoop,lasp}. For instance, Yao et al. \cite{kgcoop} adopt human knowledge (\textit{a photo of a} [\textit{class name}]) as hard prompts to guide the learning of soft prompts at the global level. However, the insufficient knowledge and inadequate guidance still lead to non-interpretable prompt learning. To address the explainability challenge of current methods, we propose \textbf{XCoOp}, a novel e\textbf{X}plainable prompt learning framework for medical image analysis via concept-guided \textbf{Co}ntext \textbf{Op}timization, which leverages medical knowledge by aligning the semantics of the images, learnable prompts, and clinical concept-driven prompts at multiple granularities, making each token of soft prompts more informative and explainable guided by clinical concepts of corresponding diseases. Furthermore, our framework addresses the lack of valuable concept annotations by eliciting knowledge from large language models and offers both visual and textual explanations for learned prompts.


We summarize our main contributions as follows: (i) We propose XCoOp, a novel explainable prompt learning framework that leverages concept-based medical knowledge at multiple granularities to make the prompts more explainable. To the best of our knowledge, this is the first work to explore addressing the lack of interpretability of prompt learning methods. (ii) We demonstrate that our method can be flexibly applied to various datasets with or without concept annotations, alleviating the requirement of human labor by eliciting medical knowledge from LLMs. (iii) Extensive experiments and explainability analyses show that our method simultaneously achieves promising performance and interpretability, highlighting the effectiveness of foundation model-enhanced XAI.


\section{Method}

\begin{figure*}[t]
\centering
\includegraphics[width=\textwidth]{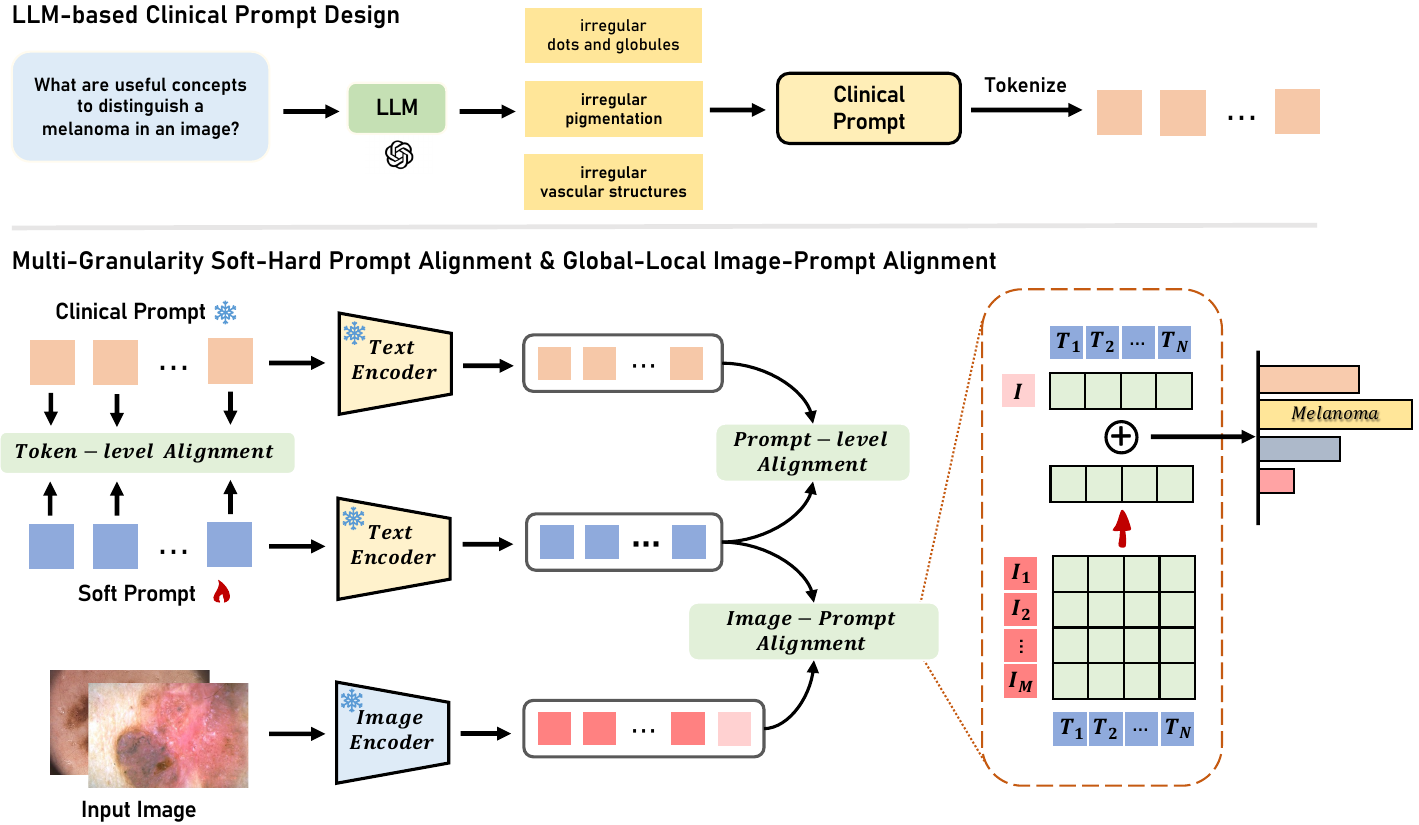}
\caption{The overall pipeline of XCoOp. The key insight of XCoOp is enhancing the informativeness and explainability of the soft prompts under the guidance of concept-based medical knowledge at multiple granularities, achieving FM-enhanced XAI.} 
\label{overview}
\end{figure*}

Fig. \ref{overview} presents the overall architecture of our explainable prompt learning framework for computer-aided diagnosis. Specifically, we initialize the soft prompts with ``\textit{a photo of a} [\textit{disease name}]'', and the clinical prompts are created based on the medical concepts (Section \ref{sec:prompt_design}). The text features of the prompts are extracted using a pre-trained text encoder, and a multi-granularity prompt alignment module is proposed to align the semantics of the soft prompts and the clinical prompts (Section \ref{sec:soft_hard}). The final disease diagnosis is performed by measuring the similarity between the text features of soft prompts and the image features at both global and local levels (Section \ref{sec:image_prompt}).


\subsection{Clinical Concept-Driven Prompt Design}
\label{sec:prompt_design}
To introduce medical knowledge into the prompt learning process, we first design disease-specific prompts using clinical concepts. Fig. \ref{overview} shows the steps of creating clinical prompts in our framework. Specifically, for medical datasets with concept annotations (e.g., Derm7pt \cite{derm7pt}, SkinCon \cite{daneshjou2022skincon}), we can easily create clinical prompts based on the labels annotated by medical experts. 
An example clinical prompt for melanoma in a dermoscopic image is \textit{a photo of {melanoma}, with irregular pigment network, dots and globules, blue-whitish veil, and vascular structures}. 
For the datasets lacking explicit concept annotations, we elicit medical knowledge from a large language model such as GPT4 \cite{achiam2023gpt} and create the corresponding clinical prompts. A sample query used to prompt the LLM is \textit{``What are the most useful visual concepts to distinguish} [\textit{disease name}] \textit{in a} \{\textit{dermoscopic image, chest X-ray, etc.}\}\textit{?''}. 

\subsection{Soft-Hard Prompt Alignment}
\label{sec:soft_hard}
To enhance the informativeness and explainability of the soft prompts by incorporating clinical semantics, we introduce a soft-hard prompt alignment module that aligns prompts at both the prompt level and token level. Prompt-level alignment facilitates the model to learn correspondences between soft prompts and clinical (hard) prompts from a global disease perspective, exploiting the intrinsic information captured by the pre-trained text encoder. The token-level alignment focuses on a more fine-grained local level. Since each token embedding of the clinical prompts is obtained by tokenizing the original concept-based prompts, the alignment enforces the soft prompts to be close to the clinical prompts in the embedding space, aiming to make each token of soft prompts more informative and explainable guiding by clinical concepts of corresponding diseases, hence enhancing the effectiveness and interpretability of the prompt learning framework. \\

\noindent\textbf{Token-level Alignment.} Given the token embeddings of soft prompts $V \in \mathbb{R}^{D \times c \times dim}$ and clinical prompts $Q\in \mathbb{R}^{D \times C \times dim}$ for different classes, we first align their embeddings at the token level via contrastive learning, where $D, C, dim$ represent the number of classes, the context length, and dimension of embedding, respectively. A probability distribution over the class labels is given by:
\begin{equation}
    P(y_d|V_d) = \frac{{\rm{exp}}({\rm{cos}}(Q_{d}, V_d)/\tau)}{\sum_{k=1}^D{\rm{exp}}({\rm{cos}}(Q_k, V_d)/\tau)},
\end{equation}

\noindent 
where $y_d$ is the binary label of class $d$, ${\rm{cos}}(\cdot,\cdot)$ is the cosine similarity, and $\tau$ is a temperature parameter. The token-level alignment objective $\mathcal{L}_T$ is optimized by minimizing the cross-entropy loss:
\begin{equation}
    \mathcal{L}_T = -\sum_{k=1}^Dy_k{\rm{log}}P(y_k|V_d).
\end{equation}



\noindent\textbf{Prompt-level Alignment.} Given the pre-trained text encoder $g(\cdot)$, we align the text features of soft prompts and clinical prompts at the global prompt level by minimizing the following objective function:
\begin{equation}
    \mathcal{L}_P = \sum_{d=1}^DCE(\frac{{\rm{exp}}({\rm{cos}}(g(Q_{d}), g(V_d)))/\tau)}{\sum_{k=1}^D {\rm{exp}}({\rm{cos}}(g(Q_k), g(V_d))/\tau)}, y_d),
\end{equation}
where $\mathcal{L}_P$ represents the prompt-level alignment loss, $CE(\cdot)$ denotes the cross-entropy loss, $g(V)$ and $g(Q)\in \mathbb{R}^{D \times dim} $ denote the output text features of soft prompts $V$ and clinical prompts $Q$, respectively. The overall objective of the soft-hard prompt alignment module $\mathcal{L}_{\rm{PPA}}$ is the average of $\mathcal{L}_T$ and $\mathcal{L}_P$.

\subsection{Global-Local Image-Prompt Alignment}
\label{sec:image_prompt}
Medical diagnosis typically hinges on various clinical symptoms observable within specific, localized regions in an image. 
Given that different clinical concepts may correspond to distinct sub-regions of a medical image, we employ a global-local image-prompt alignment module to align the medical images and clinical concept-driven prompts at multiple levels. Specifically, as illustrated in Fig.  \ref{overview}, given an image $x$ and the pre-trained image encoder of CLIP \cite{clip}, we obtain the global visual feature $p$ and a set of local features $F = \{f_1, f_2, ..., f_M\}$, where $M$ is the number of local (patch) features. The final prediction probability is computed by the matching scores of both global and local features, and the alignment can be optimized using cross-entropy loss which estimates the discrepancy between the predicted diagnosis results and the ground truth:
\begin{equation}
    \mathcal{L}{\rm{_{IPA}}} = CE[{\rm{cos}}(p,g(V_d)) + \lambda\frac{1}{M}(\sum_{m=1}^M {\rm{cos}}(f_m,g(V_d))), y_d],
\end{equation}
where $\mathcal{L}{\rm{_{IPA}}}$ represents the image-prompt alignment loss, $\lambda$ is the weight of the prediction of local features. The overall training objective is represented as $\mathcal{L} = \mathcal{L}{\rm{_{PPA}}} + \lambda'\mathcal{L}{\rm{_{IPA}}}$, where $\lambda'$ is a loss-balancing factor. The global-local image-prompt alignment module encourages the model to mimic the process wherein medical experts utilize both global and local information to diagnose disease.

\section{Experiments}

\subsection{Experimental Setup}

\noindent\textbf{Datasets:} \textbf{\textit{Derm7pt}} \cite{derm7pt} is a dermoscopic image dataset contains 1011 images with clinical concepts for melanoma skin lesions in dermatology. 
\textbf{\textit{SkinCon}} \cite{daneshjou2022skincon} is a skin disease dataset densely annotated by experts for fine-grained model debugging and analysis. 
The concepts of \textit{Derm7pt} and \textit{SkinCon} are used to design clinical prompts for these two datasets.
\textbf{\textit{Pneumonia}} \cite{pneumonia} is a public dataset for classifying pneumonia cases from normal ones.  \textbf{\textit{IU X-Ray}} \cite{iu-xray} is a chest X-ray dataset 
with 3,955 radiology reports, corresponding to 7,470 frontal and lateral images. We filter out the lateral x-ray, leaving only frontal images. \\
\noindent\textbf{Implementation Details:}
Our framework adopted the pre-trained visual (ViT-B/16) and text encoder of CLIP \cite{clip}. 
We adopted SGD \cite{sgd} optimizer with learning rate of 0.032. We used warm-up and cosine anneal as training strategies. All methods implemented in this paper adopted random crop and random flip for data augmentation. Grid search was used to select hyperparameters, we set $\tau=0.9, \lambda=0.1$. All experiments were conducted on an RTX 3090 GPU. 



\definecolor{green2}{hsb}{0.5, 0.2, 1}
\setlength{\tabcolsep}{0.7mm}  
\begin{table*}[b]  
\caption{Quantitative comparison on disease diagnosis with the state-of-the-art prompt learning methods. The performance is reported as mean$_{\rm std}$ of three random runs. Our method is highlighted in light cyan, and the best results are shown in \textbf{bold}.}
\centering  
\fontsize{8.5}{11}\selectfont  
\begin{threeparttable}  
	  
\begin{tabular}{c|cc|cc|cc|cc}  

    \toprule\hline
    \multirow{2}{*}{\bf METHOD}&
    \multicolumn{2}{c|}{\bf Derm7pt}&
    \multicolumn{2}{c|}{\bf SkinCon}&
    \multicolumn{2}{c|}{\bf Pneumonia}&
    \multicolumn{2}{c}{\bf IU X-Ray}
    \\

    &\bf AUC  &\bf ACC   &\bf AUC  &\bf ACC   &\bf AUC  &\bf ACC  &\bf AUC  &\bf ACC \cr
                

    \hline\hline 
    \textcolor{gray}{CLIP \cite{clip}} &\textcolor{gray}{50.00$_{}$}&\textcolor{gray}{69.11$_{}$} &\textcolor{gray}{39.68$_{}$} &\textcolor{gray}{70.29$_{}$ } &\textcolor{gray}{50.00$_{}$}&\textcolor{gray}{62.52$_{}$ }&\textcolor{gray}{47.90} & \textcolor{gray}{13.21}\cr
    \hline\hline
    CoOp \cite{coop} &71.76$_{0.1}$&75.19$_{0.4}$ &77.52$_{0.4}$ &75.91$_{0.7}$   &84.08$_{0.6}$&85.88$_{0.6}$ &78.45$_{1.2}$ & 71.93$_{0.7}$\cr
    CoCoOp \cite{cocoop} &70.40$_{0.4}$&77.04$_{0.7}$ &78.02$_{0.5}$&76.19$_{0.8}$  &85.96$_{0.4}$&86.06$_{0.8}$ & 76.00$_{1.6}$ & 70.63$_{0.5}$\cr  
    KgCoOp \cite{kgcoop} &69.67$_{2.7}$&73.84$_{1.4}$ & 75.33$_{0.3}$&76.95$_{0.5}$ & 80.95$_{0.3}$ & 82.64$_{0.3}$ & 75.61$_{1.2}$ & 70.74$_{1.2}$ \cr  
    LASP \cite{lasp} &75.08$_{0.6}$ & {76.20$_{1.6}$}  &{78.31$_{0.3}$}&{77.33$_{0.8}$} &91.31$_{0.1}$& 92.41$_{0.1}$ & 83.69$_{0.3}$  & 76.46$_{0.7}$  \cr   
    \hline\hline 
    \rowcolor{green2!40}\bf XCoOp & \bf{78.43$_{0.6}$} & \bf{78.82$_{1.0}$}&    \bf{81.12$_{0.3}$}&\bf{78.57$_{0.6}$} & \bf{92.85$_{0.3}$}& \bf{93.80$_{0.3}$} &\bf{84.91$_{0.6}$} &\bf{78.44$_{0.9}$}\cr

    \hline
   
\end{tabular}  
\end{threeparttable} 

\label{tab:diag_performance} 
\end{table*}

\subsection{Experimental Results}
In order to comprehensively demonstrate the competitive performance of our method in disease diagnosis, we commence by comparing with other state-of-the-art prompt learning methods on four datasets. Subsequently, we undertake intensive ablation experiments to assess the effectiveness of our method. Finally, we evaluate the explainability of our framework using multiple XAI metrics.


\newfloat{figtab}{htb}{fgtb}
\makeatletter
  \newcommand\figcaption{\def\@captype{figure}\caption}
  \newcommand\tabcaption{\def\@captype{table}\caption}
\makeatother

\begin{figtab}[t]
\begin{minipage}[b]{0.38\textwidth} 
    \centering 
    \setlength{\belowcaptionskip}{0.3cm}  
    \setlength{\tabcolsep}{5pt}
    \tabcaption{Ablation study of XCoOp on disease diagnosis (AUC [\%]). CCP, IPA, and PPA represent the clinical concept-driven prompts, image-prompt alignment, and soft-hard prompt alignment modules, respectively.}
    \label{ablation} 
    \begin{tabular}{l|c} 
        \toprule\hline
        {\bf Method} & {AUC}\cr
        \hline
        Baseline (LASP \cite{lasp}) 
        & 78.31$_{0.3}$ \\
        \hline
        CCP
        &  79.93$_{0.4}$ \\
        CCP + IPA
        &  80.46$_{0.7}$ \\
        CCP + IPA + PPA 
        &  \bf 81.12$_{0.3}$ \\
        \hline\hline
    
    \end{tabular}
\end{minipage}\hfill \quad
\begin{minipage}[t]{0.58\textwidth} 
    \centering 
    \includegraphics[width=0.9\textwidth, height=4.5cm]{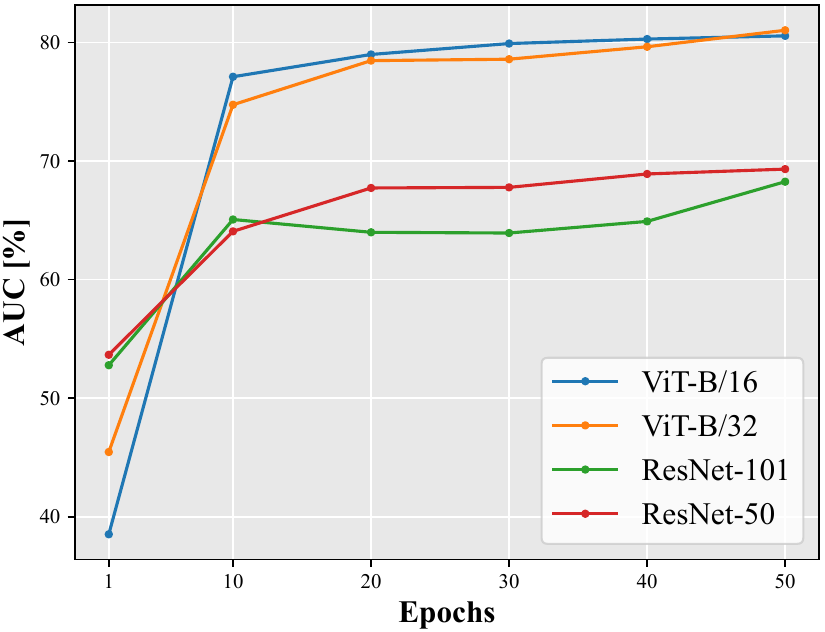} 
    \figcaption{Ablation study on the number of training epochs of XCoOp with different vision backbones.}
    \label{backbone} 
\end{minipage}
\end{figtab}

\subsubsection{Diagnosis Results.}
In Table \ref{tab:diag_performance}, we report the disease diagnosis comparison results of our method using AUROC and Accuracy on four medical image datasets. 
We include the CLIP baseline \cite{clip} without any tuning (the first row), two CoOp-based methods (CoOp \cite{coop} and CoCoOp \cite{cocoop}), and two knowledge-guided prompt learning methods (KgCoOp \cite{kgcoop} and LASP \cite{lasp}). Our method outperforms other state-of-the-art prompt learning methods with a significant margin, especially achieving $1.2\% \sim 3.4\%$ AUC and  $1.2\% \sim 2.0\%$ accuracy improvement compared to the second-best results on all considered datasets, which demonstrates that the full utilization of medical knowledge and the global-local correlations between images and prompts effectively encourages the soft prompts to learn clinical semantics, thus benefiting the performance of our model.

\subsubsection{Ablation Study.}
We conduct various ablation studies on \textit{SkinCon} to investigate the influence of different modules and settings. In Table \ref{ablation}, we assess the effectiveness of each module in our proposed framework. Specifically, we show that our method can benefit from all the components, including the clinical concept-driven prompts, the soft-hard prompt alignment, and the global-local image-prompt alignment. The last configuration of Table \ref{ablation} 
demonstrates that our method achieves the best overall performance with all designed components. To explore the influence of different numbers of training epochs and vision backbones, we conduct an ablation study and report the AUC in Fig. \ref{backbone}. 
The results show that our method can converge within around 35 epochs with different vision backbones (e.g., ViT \cite{vit}, ResNet \cite{resnet}), which demonstrates the high efficiency and robustness of our method.

\begin{figure*}[t]
\centering
\subfigure[Prompt examples.]
{
    \label{intervene} 
    \includegraphics[width=0.4\columnwidth, height=3.6cm]{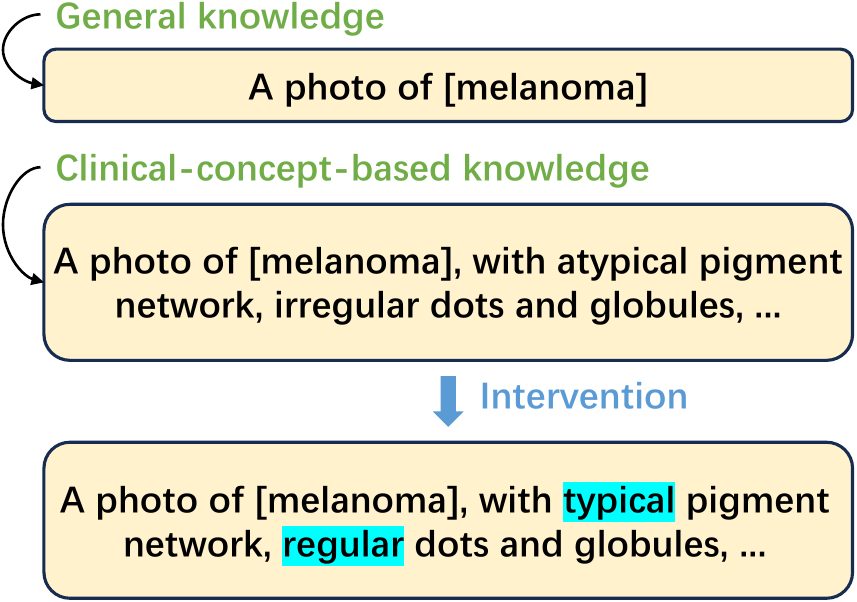} 
}
\quad\
\subfigure[Knowledge intervention.]
{   
    \label{intervene_result} 
    \includegraphics[width=0.4\columnwidth, height=3.6cm]{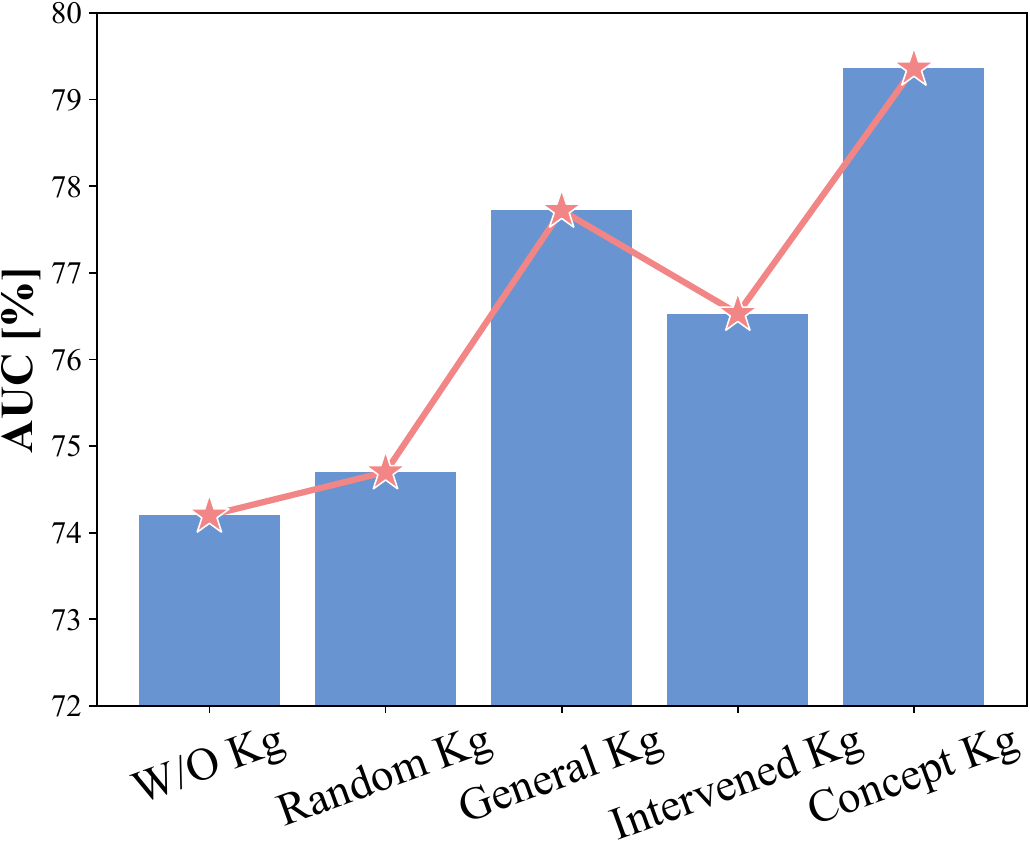} 
}

\caption{Illustration of our model's faithfulness using knowledge intervention. (a) Clarification of prompt examples based on different knowledge and intervention. (b) The results of concept-based knowledge (Kg) intervention on Derm7pt, where x-axis represents different kinds of prompts and y-axis represents the AUC [\%], respectively.} 
\label{kg_intervention}
\end{figure*}

\subsection{Analysis of Explainability}
In order to evaluate the explainability of our proposed method, we analyze our framework using multiple crucial XAI metrics in this section. Specifically, inspired by previous works \cite{guidotti2018survey,rigotti2021attention,hsiao2021roadmap,bie2024mica,jin2023guidelines}, we evaluate our framework from the perspectives of \textit{faithfulness}, \textit{understandability} and \textit{plausibility}. \\

\noindent\textbf{Faithfulness.}
\textit{Faithfulness} is defined as the degree to which the explanation reflects the model decision process and requires the explanation to be faithful to the designed model mechanism \cite{guidotti2018survey,lakkaraju2019faithful,rigotti2021attention}. In this paper, we evaluate \textit{faithfulness} by intervening the input clinical concept-driven prompts. As shown in Fig. \ref{kg_intervention}, we use five kinds of prompt settings, including without knowledge, with random knowledge (i.e., random words as clinical prompts), with general knowledge (i.e., knowledge without specific clinical concepts), with clinical-concept-based knowledge and the intervened knowledge. Specifically, we adopt \textit{Derm7pt} dataset as an example, as shown in Fig. \ref{intervene}, where intervention means modifying some of the concepts in the original clinical prompts and obtaining a new prompt. Fig. \ref{intervene_result} shows that using only random knowledge, general knowledge, or knowledge after intervention as prompts may lead to performance degradation, which demonstrates that the clinical knowledge faithfully explains the model's decisions. \\

\begin{figure*}[t]
\centering
\subfigure[Image-prompt similarity visualization.]
{
    \label{vis} 
    \includegraphics[width=0.7\columnwidth, height=3.5cm]{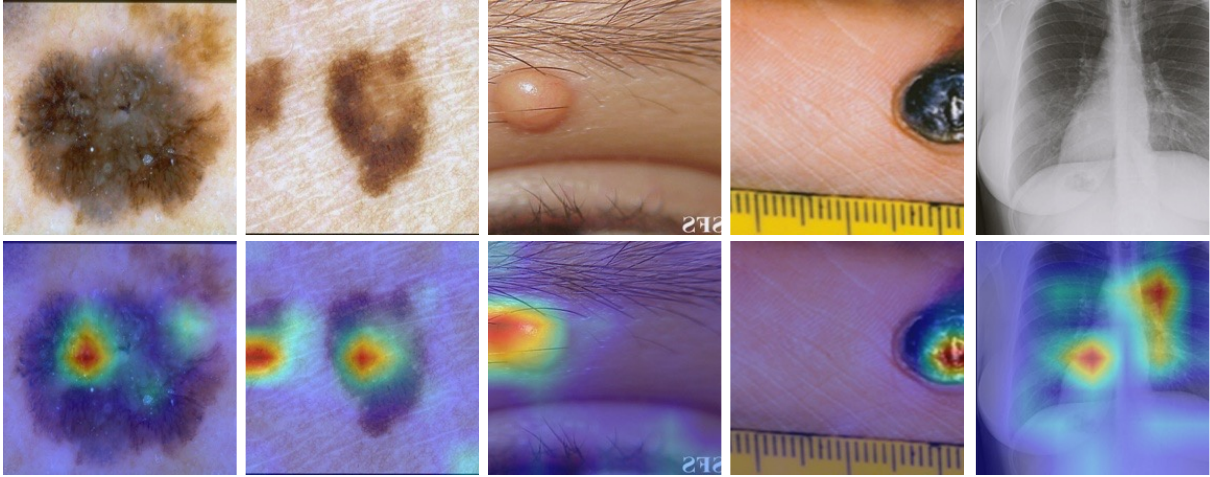} 
}
\quad\
\subfigure[t-SNE.]
{   
    \label{t_sne} 
    \includegraphics[width=0.2\columnwidth, height=3.5cm]{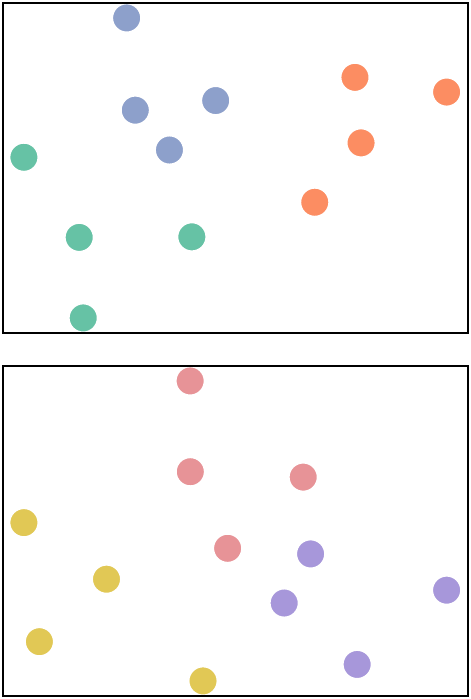} 
}
\caption{Visual explanations. (a) Visualization examples of the similarity between the images and soft prompts. (b) The t-SNE visualization of tokens of different soft prompts of SkinCon (top) and IU X-ray (bottom) datasets. Different colors represent different categories of prompts, and the number of context tokens is 4.} 
\label{visualization}
\end{figure*}

\noindent\textbf{Understandability \& Plausibility.}
\textit{Understandability} requires explanations to be easily understandable by users without requiring technical knowledge \cite{jin2023guidelines} while \textit{plausibility} refers to given domain knowledge, how believable or likely the explanation seems \cite{guidotti2018survey,carvalho2019machine}. Our framework achieves \textit{understandability} and \textit{plausibility} by offering both textual and visual explanations. Specifically, we interpret the learnable prompts by measuring the distance between the soft prompts and the hand-crafted clinical prompts. As shown in Table \ref{interpretation}, we compare the average distances with two knowledge-guided prompt learning methods \cite{kgcoop,lasp}. Our method outperforms the other methods and achieves the least distance between the learnable prompts and the clinical prompts. For visual explanation, our framework provides the similarity visualization between the medical images and the learnable prompts, as shown in Fig. \ref{vis}, where we can observe that the model focuses more on discriminative concept regions within images guided by our learned prompts. Fig. \ref{t_sne} presents the t-SNE visualization \cite{t_sne} of tokens of different soft prompts and shows that the token embeddings cluster well, demonstrating that tokens in each prompt meticulously learn the discriminative clinical semantics of the corresponding disease category. The explanations offered by our framework enhance human comprehension of the model's decision-making process by elucidating the utilized knowledge and the specific regions of focus, potentially aiding medical experts in utilizing AI models for disease diagnosis.
\setlength{\tabcolsep}{1mm}  
\begin{table*}[t]  
\caption{Quantitative comparison on prompt interpretation by measuring distances between the soft prompts and the hand-crafted clinical prompts (i.e., textual explanations). The results are reported as the average distances of different categories. Our method is highlighted in light cyan, and the best results are shown in \textbf{bold}.}
\label{interpretation}
\centering  
\fontsize{9}{10}\selectfont  

\scalebox{1}{\begin{tabularx}{\textwidth}{c|>{\centering\arraybackslash}X|>{\centering\arraybackslash}X|>{\centering\arraybackslash}X|>{\centering\arraybackslash}X|>{\centering\arraybackslash}X}
        \toprule\hline
        {\bf Method} & Derm7pt & SkinCon &Pneumonia &IU X-ray 
        & \bf Average  $\downarrow$ \cr
        \hline\hline
        KgCoOp \cite{kgcoop}
        &  2.293 & 1.475& 1.727&2.433&1.982 \\
        LASP \cite{lasp}
        &  2.936 & 3.867 & 2.270 &2.972 & 3.011\\
        \rowcolor{green2!40}\bf XCoOp 
        &  \bf 1.161 & \bf 1.139 &\bf 0.987 & \bf 1.127 & \bf 1.104 \\
        \hline

\end{tabularx}}
\end{table*}

\section{Conclusion}
In this paper, we propose XCoOp, an explainable prompt learning framework for computer-aided diagnosis, which utilizes medical knowledge by aligning the semantics of images, learnable prompts, and clinical concept-driven prompts at multiple granularities. By adopting the concept-based knowledge eliciting from foundation models to guide the soft prompt at both the token embedding level and prompt level, our method outperforms other prompt learning methods while preserving inherent interpretability with both visual and textual explanations. Extensive experiments and explainability analyses conducted on various datasets demonstrate that our method simultaneously achieves promising performance and interpretability, highlighting the effectiveness of FM-enhanced XAI.

\bibliographystyle{splncs04}
\bibliography{miccai2024}

\begin{thebibliography}{10}
\providecommand{\url}[1]{\texttt{#1}}
\providecommand{\urlprefix}{URL }
\providecommand{\doi}[1]{https://doi.org/#1}

\bibitem{achiam2023gpt}
Achiam, J., Adler, S., Agarwal, S., Ahmad, L., Akkaya, I., Aleman, F.L., Almeida, D., Altenschmidt, J., Altman, S., Anadkat, S., et~al.: Gpt-4 technical report. arXiv preprint arXiv:2303.08774  (2023)

\bibitem{alayrac2022flamingo}
Alayrac, J.B., Donahue, J., Luc, P., Miech, A., Barr, I., Hasson, Y., Lenc, K., Mensch, A., Millican, K., Reynolds, M., et~al.: Flamingo: a visual language model for few-shot learning. Advances in Neural Information Processing Systems  \textbf{35},  23716--23736 (2022)

\bibitem{bie2024mica}
Bie, Y., Luo, L., Chen, H.: Mica: Towards explainable skin lesion diagnosis via multi-level image-concept alignment. arXiv preprint arXiv:2401.08527  (2024)

\bibitem{lasp}
Bulat, A., Tzimiropoulos, G.: Lasp: Text-to-text optimization for language-aware soft prompting of vision \& language models. In: Proceedings of the IEEE/CVF Conference on Computer Vision and Pattern Recognition. pp. 23232--23241 (2023)

\bibitem{carvalho2019machine}
Carvalho, D.V., Pereira, E.M., Cardoso, J.S.: Machine learning interpretability: A survey on methods and metrics. Electronics  \textbf{8}(8), ~832 (2019)

\bibitem{daneshjou2022skincon}
Daneshjou, R., Yuksekgonul, M., Cai, Z.R., Novoa, R., Zou, J.Y.: Skincon: A skin disease dataset densely annotated by domain experts for fine-grained debugging and analysis. Advances in Neural Information Processing Systems  \textbf{35},  18157--18167 (2022)

\bibitem{iu-xray}
Demner-Fushman, D., Kohli, M.D., Rosenman, M.B., Shooshan, S.E., Rodriguez, L., Antani, S., Thoma, G.R., McDonald, C.J.: Preparing a collection of radiology examinations for distribution and retrieval. Journal of the American Medical Informatics Association  \textbf{23}(2),  304--310 (2016)

\bibitem{vit}
Dosovitskiy, A., Beyer, L., Kolesnikov, A., Weissenborn, D., Zhai, X., Unterthiner, T., Dehghani, M., Minderer, M., Heigold, G., Gelly, S., et~al.: An image is worth 16x16 words: Transformers for image recognition at scale. arXiv preprint arXiv:2010.11929  (2020)

\bibitem{guidotti2018survey}
Guidotti, R., Monreale, A., Ruggieri, S., Turini, F., Giannotti, F., Pedreschi, D.: A survey of methods for explaining black box models. ACM computing surveys (CSUR)  \textbf{51}(5),  1--42 (2018)

\bibitem{resnet}
He, K., Zhang, X., Ren, S., Sun, J.: Deep residual learning for image recognition. In: Proceedings of the IEEE conference on computer vision and pattern recognition. pp. 770--778 (2016)

\bibitem{hsiao2021roadmap}
Hsiao, J.H.w., Ngai, H.H.T., Qiu, L., Yang, Y., Cao, C.C.: Roadmap of designing cognitive metrics for explainable artificial intelligence (xai). arXiv preprint arXiv:2108.01737  (2021)

\bibitem{jia2021align}
Jia, C., Yang, Y., Xia, Y., Chen, Y.T., Parekh, Z., Pham, H., Le, Q., Sung, Y.H., Li, Z., Duerig, T.: Scaling up visual and vision-language representation learning with noisy text supervision. In: International conference on machine learning. pp. 4904--4916. PMLR (2021)

\bibitem{jin2023guidelines}
Jin, W., Li, X., Fatehi, M., Hamarneh, G.: Guidelines and evaluation of clinical explainable ai in medical image analysis. Medical Image Analysis  \textbf{84},  102684 (2023)

\bibitem{derm7pt}
Kawahara, J., Daneshvar, S., Argenziano, G., Hamarneh, G.: Seven-point checklist and skin lesion classification using multitask multimodal neural nets. IEEE journal of biomedical and health informatics  \textbf{23}(2),  538--546 (2018)

\bibitem{pneumonia}
Kermany, D.S., Goldbaum, M., Cai, W., Valentim, C.C., Liang, H., Baxter, S.L., McKeown, A., Yang, G., Wu, X., Yan, F., et~al.: Identifying medical diagnoses and treatable diseases by image-based deep learning. cell  \textbf{172}(5),  1122--1131 (2018)

\bibitem{lakkaraju2019faithful}
Lakkaraju, H., Kamar, E., Caruana, R., Leskovec, J.: Faithful and customizable explanations of black box models. In: Proceedings of the 2019 AAAI/ACM Conference on AI, Ethics, and Society. pp. 131--138 (2019)

\bibitem{li2022blip}
Li, J., Li, D., Xiong, C., Hoi, S.: Blip: Bootstrapping language-image pre-training for unified vision-language understanding and generation. In: International Conference on Machine Learning. pp. 12888--12900. PMLR (2022)

\bibitem{seg2}
Lin, Y., Nie, D., Liu, Y., Yang, M., Zhang, D., Wen, X.: Multi-target domain adaptation with prompt learning for medical image segmentation. In: International Conference on Medical Image Computing and Computer-Assisted Intervention. pp. 717--727. Springer (2023)

\bibitem{lipton2017doctor}
Lipton, Z.C.: The doctor just won't accept that! arXiv preprint arXiv:1711.08037  (2017)

\bibitem{segment}
L{\"u}ddecke, T., Ecker, A.: Image segmentation using text and image prompts. In: Proceedings of the IEEE/CVF Conference on Computer Vision and Pattern Recognition. pp. 7086--7096 (2022)

\bibitem{t_sne}
Van~der Maaten, L., Hinton, G.: Visualizing data using t-sne. Journal of machine learning research  \textbf{9}(11) (2008)

\bibitem{clip}
Radford, A., Kim, J.W., Hallacy, C., Ramesh, A., Goh, G., Agarwal, S., Sastry, G., Askell, A., Mishkin, P., Clark, J., et~al.: Learning transferable visual models from natural language supervision. In: International conference on machine learning. pp. 8748--8763. PMLR (2021)

\bibitem{rigotti2021attention}
Rigotti, M., Miksovic, C., Giurgiu, I., Gschwind, T., Scotton, P.: Attention-based interpretability with concept transformers. In: International Conference on Learning Representations (2021)

\bibitem{sgd}
Robbins, H., Monro, S.: A stochastic approximation method. The annals of mathematical statistics pp. 400--407 (1951)

\bibitem{rudin2019stop}
Rudin, C.: Stop explaining black box machine learning models for high stakes decisions and use interpretable models instead. Nature machine intelligence  \textbf{1}(5),  206--215 (2019)

\bibitem{kgcoop}
Yao, H., Zhang, R., Xu, C.: Visual-language prompt tuning with knowledge-guided context optimization. In: Proceedings of the IEEE/CVF Conference on Computer Vision and Pattern Recognition. pp. 6757--6767 (2023)

\bibitem{yu2205coca}
Yu, J., Wang, Z., Vasudevan, V., Yeung, L., Seyedhosseini, M., Wu, Y.: Coca: Contrastive captioners are image-text foundation models. arxiv 2022. arXiv preprint arXiv:2205.01917

\bibitem{cocoop}
Zhou, K., Yang, J., Loy, C.C., Liu, Z.: Conditional prompt learning for vision-language models. In: Proceedings of the IEEE/CVF Conference on Computer Vision and Pattern Recognition. pp. 16816--16825 (2022)

\bibitem{coop}
Zhou, K., Yang, J., Loy, C.C., Liu, Z.: Learning to prompt for vision-language models. International Journal of Computer Vision  \textbf{130}(9),  2337--2348 (2022)

\end{thebibliography}

%





\end{document}